\documentclass[12pt]{article}

\usepackage{url}

\usepackage{times}
\usepackage{latexsym}

\usepackage{amsmath,amssymb,amsthm}

\usepackage{hyperref}
\hypersetup{
    colorlinks = true,
    citecolor = {blue},
    linkcolor = {blue},
}
\usepackage{natbib}
\usepackage{booktabs}

\usepackage[linesnumbered, ruled, vlined]{algorithm2e}

\usepackage{graphicx}

\usepackage{xcolor}

\newtheorem{remark}{Remark}
\theoremstyle{remark}

\renewcommand{\P}{\mathbb{P}}
\newcommand{\R}{\mathbb{R}}
\newcommand{\dirich}{\mathrm{Dir}}
\newcommand{\mult}{\mathrm{Multinomial}}
\newcommand{\cusum}{\mathrm{CU}}
\newcommand{\lr}{\mathrm{LR}}
\newcommand{\hK}{\widehat{K}}
\newcommand{\hPhi}{\widehat{\Phi}}
\newcommand{\full}{\mathrm{full}}
\newcommand{\ttmc}{\mathrm{TTMC}}
\newcommand{\EstimateTopicCounts}{\textsc{EstimateTopicCounts}}
\newcommand{\CalculateThresholds}{\textsc{CalculateThresholds}}
\newcommand{\mwbs}{\textsc{MWBS}}
\newcommand{\EstimateChangepoints}{\textsc{EstimateChangepoints}}
\newcommand{\LearnedTopicModel}{\texttt{LearnedTopicModel}}
\newcommand{\scores}{\texttt{scores}}
\newcommand{\intervals}{\texttt{intervals}}
\newcommand{\Threshold}{\texttt{Threshold}}
\newcommand{\changepoints}{\texttt{changepoints}}
\newcommand{\preChangeScores}{\texttt{preChangeScores}}
\newcommand{\postChangeScores}{\texttt{postChangeScores}}
\newcommand{\preChangeIntervals}{\texttt{preChangeIntervals}}
\newcommand{\postChangeIntervals}{\texttt{postChangeIntervals}}
\newcommand{\statistic}{\texttt{statistic}}

\newcommand*\mathinhead[2]{\texorpdfstring{$\boldsymbol{#1}$}{#2}}

\newcommand{\blind}{0}

\usepackage{microtype}
\usepackage{cmbright}

\begin{document}

\def\spacingset#1{\renewcommand{\baselinestretch}%
{#1}\small\normalsize} \spacingset{1}


\if0\blind
{
  \title{\bf Changepoint Analysis of Topic Proportions in Temporal Text Data}
  \author{Avinandan Bose \\
    Computer Science and Engineering \\
    Indian Institute of Technology Kanpur \\
    and \\
    Soumendu Sundar Mukherjee\thanks{
    Supported by an INSPIRE Faculty Fellowship from the Department of Science and Technology, Government of India}\hspace{.2cm} \\
    Interdisciplinary Statistical Research Unit \\
    Indian Statistical Institute Kolkata}
    \date{}
  \maketitle
} \fi

\if1\blind
{
  \bigskip
  \bigskip
  \bigskip
  \begin{center}
    {\LARGE\bf Title}
\end{center}
  \medskip
} \fi

\bigskip
\begin{abstract}
    Changepoint analysis deals with unsupervised detection and/or estimation of time-points in time-series data, when the distribution generating the data changes. In this article, we consider \emph{offline} changepoint detection in the context of large scale textual data. We build a specialised temporal topic model with provisions for changepoints in the distribution of topic proportions. As full likelihood based inference in this model is computationally intractable, we develop a computationally tractable approximate inference procedure. More specifically, we use sample splitting to estimate topic polytopes first and then apply a likelihood ratio statistic together with a modified version of the wild binary segmentation algorithm of \cite{fryzlewicz2014wild}. Our methodology facilitates automated detection of structural changes in large corpora without the need of manual processing by domain experts. As changepoints under our model correspond to changes in topic structure, the estimated changepoints are often highly interpretable as marking the surge or decline in popularity of a fashionable topic. We apply our procedure on two large datasets: (i) a corpus of English literature from the period 1800-1922 \citep{englit2015}; (ii) abstracts from the High Energy Physics arXiv repository~\citep{clement2019arxiv}. We obtain some historically well-known changepoints and discover some new ones.
\end{abstract}

\noindent%
\vfill

\newpage
\spacingset{1.2} 

\section{Introduction} \label{sec:intro}
Changepoint analysis, which consists of detection and/or estimation of time-points at which the data-generating distribution changes, is an old and important topic in Statistics going back to the early days of statistical quality control (SQC) \citep{page1954continuous,page1957problems,girshick1952bayes}. There are two types of changepoint problems: (a) offline, where the whole time-series is available to the statistician; (b) online, where data are still arriving at the time of analysis. We will be concerned with the offline problem in this article.

There are $M \ge 0$ changepoints in (discrete) time-indexed data $X_1, \ldots, X_T$ if  there exist epochs $\tau_0 := 0 < \tau_1 < \cdots < \tau_M < \tau_{M + 1} := T$ such that
\[
    X_t \sim \P^{(k)} \text{ when } \tau_{k} < t \leq \tau_{k+1}, 0 \le k \le M,
\]
where $\P^{(k)}$'s are distinct probability distributions. The goal of changepoint detection is to test if $M > 0$. On the other hand, the goal of changepoint estimation is to estimate the unknown parameters $M, \tau_1, \ldots, \tau_M$. Here the $X_t$'s can be any type of random objects---random variables, random vectors, random functions, etc.

There is a huge body of literature on the univariate changepoint problem. An excellent treatment can be found in the book \cite{brodsky2013nonparametric}.

Some notable works on the multivariate version of the problem are \cite{zhang2010detecting,siegmund2011detecting,srivastava1986likelihood,james1992asymptotic} in parametric settings, and \cite{harchaoui2009kernel,lung2011homogeneity,chen2015} in nonparametric settings.

Although a lot of work has been done on changepoint detection for Euclidean time-series data, there is comparatively little existing work on more structured data. In this article, we concern ourselves with (offline) changepoint analysis of textual data. To the best of our knowledge the only works in this direction are \cite{kim2015reading, partovibayesian2015, wang2018real, sun2020topic}. Of these \cite{kim2015reading} employ a Bayesian online method of \cite{adams2007bayesian}. \cite{wang2018real} also consider the online problem and employ a temporal topic model. In the offline context, \cite{partovibayesian2015} develops a method based on a multinomial model of word frequencies. Their method is not, however, scalable to large text corpora.

We should mention here that general continuously evolving dynamic topic models have been studied before, although not from the point of view of changepoint analysis. Some notable works are \cite{blei2006dynamic, iwata2010online}. Another related field is topic segmentation \citep{eisenstein2008bayesian,du2013topic,blei2001topic,reynar1998topic,purver2011topic}, where one wants to demarcate chunks of text that belong to particular topics. Our goal, however is to compare distributions of topics in a sequence of documents which is a \textit{global} detection/estimation problem unlike topic segmentation.

\subsection{Our contributions}
In this article, we build a novel topic model with provisions for changepoints (Section~\ref{sec:model}) and develop offline algorithms for detecting and estimating changepoints based on this model (Section~\ref{sec:est}). We also consider an alternative approach of using Latent Semantic Analysis (LSA) to obtain low-dimensional representations of the documents in our corpus and then apply off-the-shelf multivariate changepoint detection algorithms on those low-dimensional representations. We also compare our method with some other baseline algorithms which use off-the-shelf univariate changepoint algorithms on cosine-similarity scores.

The main advantages of our approach are as follows: (i) the algorithms we use are quite fast even on very large corpora; (ii) the approach is quite robust against false positives unlike other baselines; (iii) the calculated statistics have relatively low variance (compared to, e.g., the LSA based approach mentioned earlier); (iv) as our approach is based on topic modelling, we often get natural interpretations for the detected changepoints (see, for instance, Section~\ref{sec:real-data}).

Since good estimation of topics needs a fairly large number of documents and the number of words per document should be decent as well, our topic-modelling based approach is not suitable for data sets where the documents are too short (e.g., tweets), or where the observed time-series is extremely short (which is anyway challenging for changepoint analysis).

Our method facilitates automated detection of structural changes in large corpora without the need of manual processing by domain experts. This could be useful in demarcating eras that a particular domain may have had over a long period of time without much domain specific knowledge. The interpretability of our method in terms of topics can serve to explain these changes. We demonstrate this using two data sets in Section~\ref{sec:real-data}: (i) a corpus of English literature from the period 1800-1922 \citep{englit2015}; (ii) abstracts from the High Energy Physics arXiv repository~\citep{clement2019arxiv}. We obtain some historically well-known changepoints and discover some new ones.

Even though there are several baseline text data sets from different domains, there are not many that span long time periods with good coverage. We hope that the availability of automated methods like ours will lead to the creation and standardization of such data sets.

\section{Preliminaries} \label{sec:prelim}
We first describe a couple of popular test statistics for detecting a single changepoint. These can be used with the wild binary segmentation (WBS) (Section~\ref{sec:wbs} below) to detect multiple changes.

Suppose that we have data $X_s, \ldots, X_e$ with a single changepoint at $s \le \tau < e$ and that the change happens in the parameter $\theta$ of the density $f_\theta(x)$ of the data-points, i.e.
\[
    X_t \sim \begin{cases}
        f_{\theta_1}(\cdot) & \text{ if } s \le t \le \tau, \\
        f_{\theta_2}(\cdot) & \text{ if } \tau < t \le e.
    \end{cases}
\]
The likelihood ratio (LR) statistic for testing the hypothesis $H_0 : \tau = t$ is given by
\begin{equation}\label{eq:lr}
    T_{s, e}^{(\lr)}(t) = \frac{\max\limits_{\theta_1, \theta_2} \prod\limits_{s \le a \le t}f_{\theta_1}(X_a) \prod\limits_{t < a \le e} f_{\theta_2}(X_a)}{\max\limits_{\theta} \prod\limits_{s \le a \le e} f_\theta(X_a)}.
\end{equation}
The likelihood ratio statistic for testing if there is a change is given by
\begin{equation*}
    T_{s, e}^{(\lr)} = \max_{s \le t < e} T_{s, e}^{(\lr)}(t).
\end{equation*}
In fact, the maximizing $t$ above is the MLE of $\tau$. We will be using the LR statistic as part of our approximate inference procedure (see Section~\ref{sec:est}).

Another highly popular statistic in the changepoint literature is the cumulative sum (CUSUM) statistic \citep{page1954continuous}. In a location model, this statistic is given by 
\begin{equation}\label{eq:cusum}
    T_{s, e}^{(\cusum)} = \max_{s \le t < e} \bigg[\frac{t}{e - s + 1}\bigg(1 - \frac{t}{e - s + 1}\bigg)\bigg]^{1/2} \bigg|\frac{1}{t-s+1}\sum_{a = s}^t X_a - \frac{1}{e - t}\sum_{a = t + 1}^e X_a \bigg|.
\end{equation}
The idea behind this statistic is that at $t = \tau$ the pre-$t$ and post-$t$ averages will be the most far-apart, essentially matching the signal strength $\|\theta_1 - \theta_2\|$\footnote{Here $\|\theta_1 - \theta_2\|$ denotes some measure of distance between $\theta_1$ and $\theta_2$.} up to small random fluctuations. The maximizing $t$ above gives a bona fide estimate of $\tau$. In fact, this is the maximum likelihood estimator (MLE) of $\tau$ when $f_{\theta_i}$ is the density of $\mathcal{N}(\theta_i, 1), i = 1, 2$.

\subsection{Binary segmentation}\label{sec:wbs}
Binary segmentation (BS) is a popular algorithm for detecting and estimating multiple changepoints. The idea is to compute some changepoint statistic such as $T_{s, e}^{(\cusum)}$ or $T_{s, e}^{(\lr)}$ for the (discrete) interval $[s, e] := \{s, s + 1, \ldots, e - 1, e\}$ and record the maximizing $t \in [s, e - 1]$, say $t_0$. If the value of the chosen statistic does not exceed a threshold $\zeta$ (which is essentially a quantile of the statistic under the null model of no change), then the algorithm stops, declaring that there are no changepoints. Otherwise, one declares $t_0$ to be a changepoint, and then performs the same computation again on the two sub-intervals $[s, t_0]$ and $[t_0 + 1, e]$. The name binary segmentation derives from this recursive bipartitioning---at each stage, we potentially find a changepoint and divide the interval under investigation into two smaller intervals for further investigation. The algorithm stops when the lengths of the intervals falls below some pre-chosen minimum interval-length. In the end, one gets a set of changepoints.

In our approach, we use key ideas from the so-called \textit{wild binary segmentation} (WBS) algorithm of \cite{fryzlewicz2014wild} which greatly improves over standard binary segmentation.

In WBS, instead of the full interval $[s, e]$, one works with many subintervals of $[s, e]$ of random lengths. To elaborate, suppose that we want to look for change in an interval $[s, e] \subseteq [1, T]$. Let $S$ be a set of $M$ random subintervals $[s_m, e_m]$, $m = 1, \ldots,M$, whose start and end points have been drawn independently and with replacement from the set $\{1, \ldots, T\}$. Now, one constructs the subintervals $I_m = [s, e] \cap [s_m, e_m]$. On each of these subintervals, one computes a changepoint statistic $T_m$ such as the CUSUM statistic or the likelihood ratio statistic, etc. and notes for each the corresponding maximizer $t_m$ in $I_m$. One then computes the maximum of all these $T_m$. Denoting by $m_0$ the maximizing index, one declares $t_{m_0}$ as an estimated changepoint, provided $\max T_{m}$ is bigger than a certain threshold value which is estimated under a null model of no changes.

This localisation of the changepoint statistic works much better than the global approach of standard binary segmentation, especially in cases where multiple changepoints exist in $[s, e]$ that are close. The global statistic over $[s, e]$ then is muddled by the interference of signals from close changepoints (see, e.g., \cite{fryzlewicz2014wild} for a thorough discussion). By zooming into intervals containing a single changepoint, one greatly enhances the discriminating power of the chosen changepoint statistic. Finally taking a maximum over all the computed statistics churns out the changepoint with the strongest signal.

Anyway, given that $t_{m_0}$ is found to be a changepoint, one segments the original interval $[s, e]$ into two subintervals $[s, t_{m_0}]$ and $[t_{m_0}, e]$, and then applies the same algorithm on both of these subintervals. In the end of this recursive procedure, one gets hold of a set $\mathcal{C} = \{ t_{m_0}, \ldots, t_{m_\ell} \}$ of estimated changepoints. Of course, if no changepoint is found in the first step, then $\mathcal{C} = \emptyset$.

\subsection{Topic models}
Topic models, and, in particular, \textit{latent Dirichlet allocation} (LDA) \citep{blei2003latent} will be crucial to our approach. Loosely speaking, topics are collections of words (with probability scores) that represent a concept, and hence they provide useful and interpretable low-dimensional (in the sense that a document usually has only a handful of topics) representations of textual data. We describe the novel temporal topic model we use in Section~\ref{sec:model}. 

\subsection{Latent semantic analysis}\label{sec:lsa}
Topic models are one way of obtaining low-dimensional representations of textual data. There is also a model-free approach for obtaining low-dimensional (but not very interpretable) representations of textual data called \emph{latent semantic analysis} (LSA) (see, e.g., \cite{landauer1998introduction}), inspired by principal component analysis (PCA) for multivariate data. LSA goes by obtaining first the singular value decomposition of the term-document matrix that contains how many times a particular term appears in a given document, and then obtaining a low-rank approximation of that matrix by only keeping the top singular values, say $k$ many. One gets a $k$-dimensional representation of each document in the corpus by projecting the term-frequency vector of the document onto the $k$ top left-singular vectors. 

\section{Related Work} \label{sec:related-works}
\cite{partovibayesian2015} and \cite{kim2015reading} both use the Bayesian online changepoint detection (BO-CPD) method proposed by \cite{adams2007bayesian} to detect changepoints. The time series is divided into segments of contiguous documents, with the objective of maximising the joint probability of segment sizes conditioned on the documents seen so far. Enumerating this for all possible combinations is combinatorially expensive, making this approach suitable only for time series with very few documents. \cite{partovibayesian2015} focuses on the tasks of detecting sentiment change in tweets, and changes in authorship. \cite{kim2015reading} formulate their problem in terms of a generative model for the probability of the text stream given the sizes of the contiguous partitions of the stream induced by the changepoints.

\cite{wang2018real} also focus on detecting changepoints in an online setting. Unlike \cite{partovibayesian2015} and \cite{kim2015reading} who rely on simple scores like TF-IDF to model documents, \cite{wang2018real} use a topic model. After fitting a topic model on the data stream, they shortlist a few top words, and represent each document in terms of the popularity of the shortlisted top words in it. Then on these representations, they use (i) cosine similarity and (ii) Jenson-Shannon divergence between successive documents, and on these computed scores use standard changepoint detection methods like BCP, ECP, OCP and OCP+ \citep{smith1975bayesian, erdman2008bcp}.

Compared to previous works where document were represented by TF-IDF scores or top word popularity, we represent documents in our temporal topic model on a common probability simplex whose corners represent different topics, and look for changes in these representations over time. 

The methods of \cite{partovibayesian2015} and \cite{kim2015reading} are difficult to scale up to large datasets because of the computationally expensive BO-CPD algorithm of \cite{adams2007bayesian}. While the method of \cite{wang2018real} is scalable, top word popularity is more suited to cases where the documents are very short (e.g., tweets) and a few words characterise a document well. \cite{partovibayesian2015} applied his method on the authorship change detection task but the text stream he used was very small, and individual documents were small letters by different presidents of United States. In contrast, our approach can handle much larger documents---for instance, in one of our real data examples, the documents are novels by several different authors, which have orders of magnitude more words and are collected over more than a century.

\section{A temporal topic model with changepoints} \label{sec:model}
Consider a vocabulary $\mathcal{V}$ consisting of $V$ words. Each document is a \textit{bag of words} from $\mathcal{V}$. In total we have $T$ documents indexed by time. Let 
\[
    \Delta_m := \{(p_1, \ldots, p_{m + 1})^\top \in \R^{m + 1} \mid p_i \ge 0 \text{ for all } 1 \le i \le m + 1, \sum_{i = 1}^{m + 1} p_i = 1 \}
\]
denote the $m$-dimensional probability simplex.
 
Suppose that we have a corpus of $d$ documents. Let $W_d = \{w_{dn} \mid 1 \leq n \leq N_d\}$ be the $d$-th document containing $N_d$ words, where $w_{dn}$ denotes the $n$-th word in $W_d$.

\begin{figure}[!t]
\centering
\includegraphics[width=0.6\textwidth]{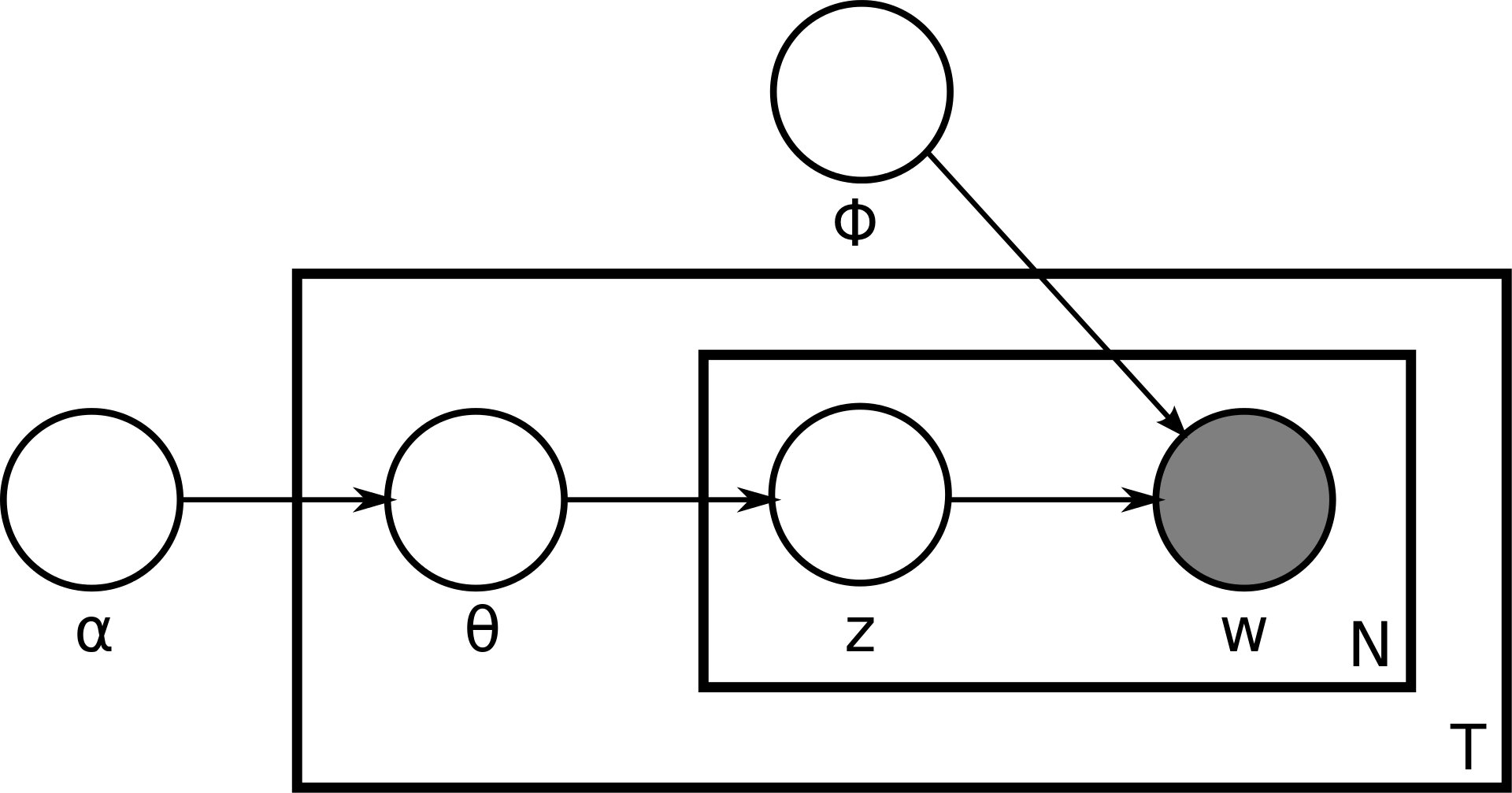}
\caption{A graphical model representation of LDA.}
\label{fig:lda}
\end{figure}

We now recall the concept of topic models. Topics are probability distributions over the vocabulary $\mathcal{V}$. Suppose that we have $K$ topics $\phi_1, \ldots, \phi_K \in \Delta_{V - 1}$. In topic models, each document is thought of as a mixture of topics in the following sense. Associated to the $d$-th document is a $K$-dimensional probability vector $\theta_d \in \Delta_{K - 1}$. For the word $w_{dn}$, a topic $z_{dn} \sim \mult(1; {\theta_d})$ is sampled first, and then $w_{dn}$ is sampled from $\mathcal{V}$ with distribution $\phi_{z_{dn}}$, i.e. $w_{dn} \sim \mult(1; \phi_{z_{dn}})$. Further, in LDA, the topic proportion vectors are assumed to be independent $\dirich(\alpha)$. See Figure~\ref{fig:lda} for a graphical model representation of LDA.

\begin{figure}[!t]
\centering
\includegraphics[width=\textwidth]{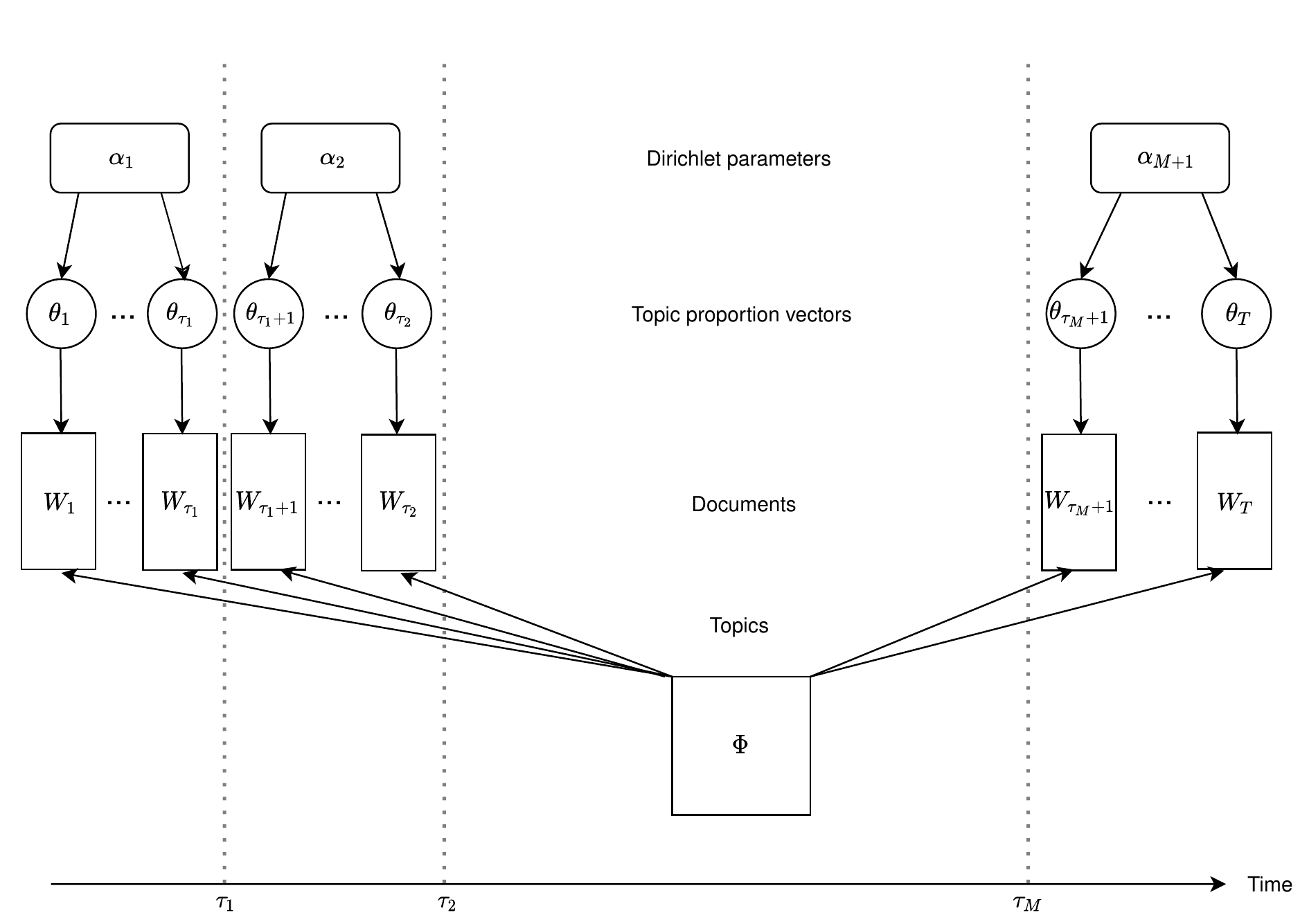}
\caption{A schematic diagram of the proposed changepoint model.}
\label{fig:model}
\end{figure}

Now we describe our \emph{temporal topic model with changepoints} for a corpus of time-indexed documents $W = (W_d)_{1 \le d \le T}$. We assume that there are $K$ latent topics $\phi_1, \ldots, \phi_K$ which remain fixed over time. Let $\Phi = [\phi_1 : \cdots : \phi_K]_{V \times K}$ denote the matrix whose columns are these topics. Let $\Theta$ = $(\theta_{d})_{1 \leq d \leq T}$ denote the sequence of topic proportion vectors. We assume $\theta_d$ to be (independent) Dirichlet distributed, the parameter of which changes over time. More precisely, we have $M$ changepoints $1 < \tau_1 < \cdots < \tau_M < T$ and associated Dirichlet parameters $\alpha_1, \ldots, \alpha_{M + 1} \in \R^K$ such that
\begin{equation}\label{eq:cpd_model}
    \theta_d \sim \dirich(\alpha_i) \text{ if } \tau_{i - 1} < d \le \tau_{i}, 1 \le i \le M + 1.
\end{equation}
Here $\tau_0 \equiv 0, \tau_{M + 1} \equiv T$. See Figure~\ref{fig:model} for a schematic diagram of this model. Note that according to the model, the documents in $W$ are independent. The temporal dependence comes only through the piecewise constant nature of the Dirichlet parameters. We will denote this model by the notation $\ttmc(T, K, \Phi, M, \tau_1, \ldots, \tau_M, \alpha_1, \ldots, \alpha_{M + 1})$. The notation
\[
    W \sim \ttmc(T, K, \Phi, M, \tau_1, \ldots, \tau_M, \alpha_1, \ldots, \alpha_{M + 1})
\]
will mean that the (time-indexed) documents in $W$ are generated according to the $\ttmc$ model on the right-hand side above.

At a first glance, it may seem restrictive to assume that the topics do not change over time. Our point of view is to have a large collection of all potential topics and then represent a document as a sparse mixture over these topics. This point of view has a major practical advantage---we need to estimate the topics only once.

\section{Estimation procedure} \label{sec:est}
\begin{figure}[!t]
\centering
\includegraphics[width=\textwidth]{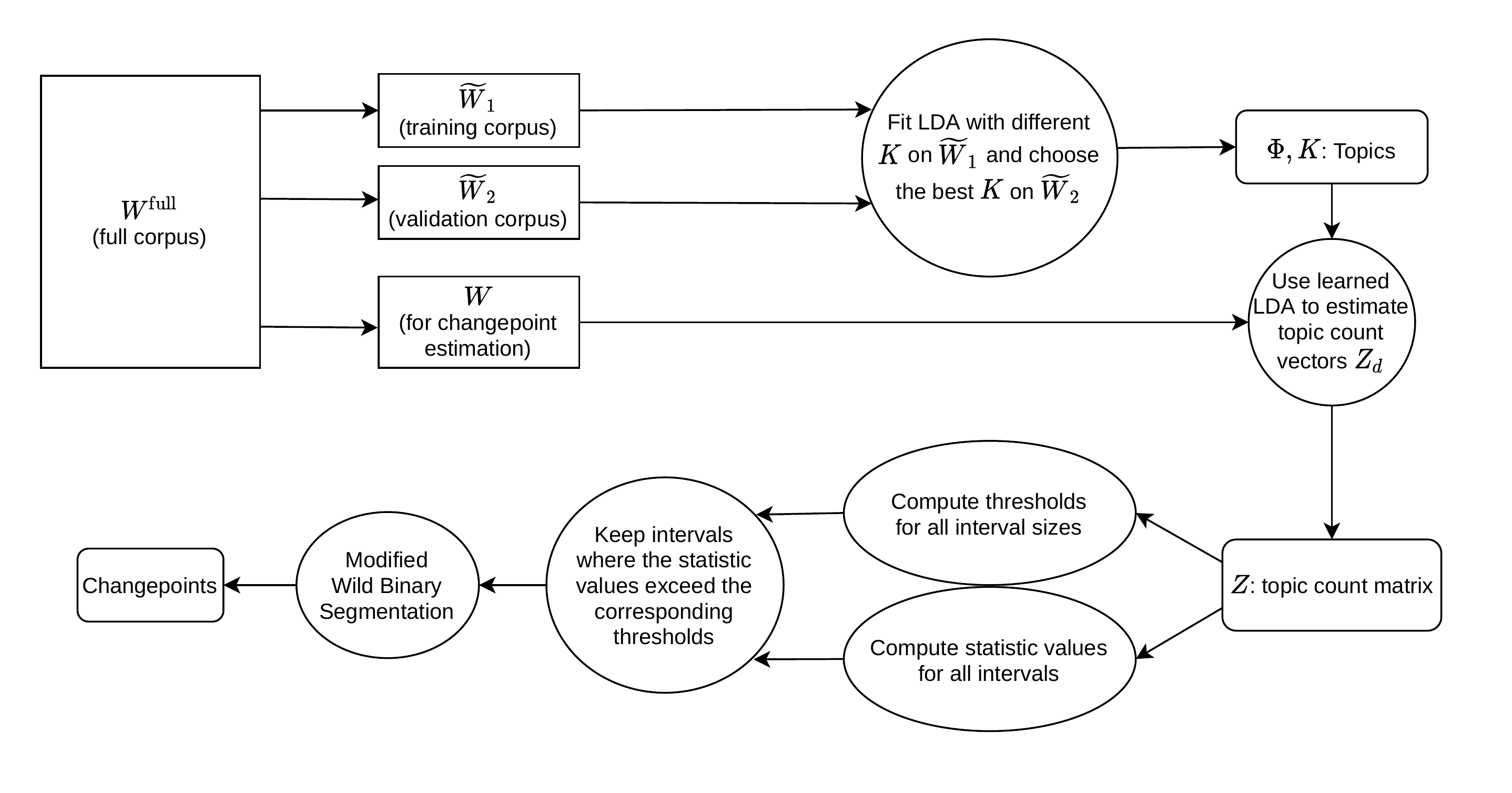}
\caption{A flowchart of our changepoint estimation procedure.}
\label{fig:method_flowchart}
\end{figure}

We now describe our estimation procedure in detail. A flowchart of the full procedure is given in Figure~\ref{fig:method_flowchart}. The first step is to estimate $K$ and $\Phi$. This is done by an application of LDA on a part of the available data. After this changepoint analysis is done on the remaining part via an application of a modified WBS-type algorithm based on an LR-type statistic.

Let us denote the full corpus by $W^{\full} = (W^{\full}_d)_{1\le d \le T_{\full}}$. We assume that
\[
W^{\full} \sim \ttmc(T^{\full}, K, \Phi, M, \tau_1^{\full}, \ldots, \tau_M^{\full}, \alpha_1, \ldots, \alpha_{M + 1}).
\]
Let $\Delta = \min_{0 \le i \le M} (\tau_{i + 1}^{\full} - \tau_i^{\full})$ denote the minimum separation between consecutive changepoints (including the left and right boundaries). We assume that $\Delta$ is reasonably large. We split $W^{\full}$ into three parts $\widetilde{W}_{1}, \widetilde{W}_{2}$ and $W$ such that 
\begin{enumerate}
    \item[(a)] each part has the same temporal order and the same number of changepoints as the full corpus;
    \item[(b)] the changepoints in each part are within a small distance $\delta$ from the corresponding changepoints in $W^{\full}$, where $\delta$ is small compared to $\Delta$.
\end{enumerate}
Such a split ensures that each part has a fair representation of documents over the entire time-series. Thus any one of them can be used to learn a good representation of the entire corpus.

One simple way of doing this, and one that we use, is to split $W^{\full}$ into three parts depending on whether a document is in the $3m$-th, $(3m + 1)$-th or $(3m + 2)$-th position:
\begin{align*}
    \widetilde{W}_1 &= \{W_d \mid d = 3m \text{ for some } m \ge 0\}, \\
    \widetilde{W}_2 &= \{W_d \mid d = 3m + 1\text{ for some } m \ge 0\}, \\
    W &= \{W_d \mid d = 3m + 2 \text{ for some } m \ge 0\}.
\end{align*}
Note that for this particular split, condition (b) is satisfied with $\delta = 2$.

We will be using $\widetilde{W} = \widetilde{W}_1 \cup \widetilde{W}_2$ to estimate $K$ and $\Phi$, and then $W$ for changepoint analysis. Note that because of condition (b), if we have good estimates of the changepoints in $W$, then these will also be good estimates of the changepoints in $W^{\full}$. 

\begin{remark}
Although we use a specific three-way split, depending on the size of the corpus, one may want to use a three-way split with different sizes, keeping more documents in the $W$-part. E.g., 
\begin{align*}
    \widetilde{W}_1 &= \{W_d \mid d = 4m + 1 \text{ for some } m \ge 0\}, \\
    \widetilde{W}_2 &= \{W_d \mid d = 4m + 3 \text{ for some } m \ge 0\}, \\
    W &= \{W_d \mid d = 4m \text{ or } 4m + 2 \text{ for some } m \ge 0\}.
\end{align*}
For this split, condition (b) is satisfied with $\delta = 1$ for $W$, and with $\delta = 3$ for $\widetilde{W}_1$ and $\widetilde{W}_2$.

Another alternative, especially when the corpus size is small, would be to take a two-way split $W^{\full} = \widetilde{W} \cup W$, use cross-validation on $\widetilde{W}$ to estimate $K$ and $\Phi$, and use $W$ for subsequent changepoint analysis.
\end{remark}

\subsection{Step 1: Estimation of \mathinhead{K}{K} and \mathinhead{\Phi}{Phi}} \label{sec:step1}
We fit LDA models with varying number of topics to $\widetilde{W}_1$, and select the model (and the corresponding estimated $\Phi$) with the best performance score on $\widetilde{W}_2$, using metrics like log-perplexity\footnote{Perplexity refers to the log-averaged inverse probability on unseen data \citep{klakow2002testing}.}. Other metrics like the ones discussed in \cite{zhao2015heuristic,greene2014many,arun2010finding} can also be used. Henceforth, we will be working with $W$ only, along with the estimated $K$ and $\Phi$. Let us denote these estimates by $\hK$ and $\hPhi$, respectively. As $W$ is independent of $\widetilde{W} = \widetilde{W}_1 \cup \widetilde{W}_2$ according to our model, the use of these estimates does not introduce any bias in the subsequent stages of our method. 

\subsection{Step 2: Estimation of latent variables} \label{sec:step2}
Suppose that $W$ has $T$ documents $W_{d_1}^{\full}, \ldots, W_{d_T}^{\full}$ and changepoints at locations $d_{\tau_1}, \ldots, d_{\tau_M}$ (recall that $|d_{\tau_i} - \tau_i^{\full}| \le \delta$ for all $1 \le i \le M$). We shall use the shorthands $W_i = W_{d_i}^{\full}$ and $\tau_i = d_{\tau_i}$. Then
\[
    W = (W_d)_{1 \le d \le T} \sim \ttmc(T, K, \Phi, M, \tau_1, \ldots, \tau_M, \alpha_1, \ldots, \alpha_{M + 1}).
\]
Let $Z_d$ denote the $K$-dimensional topic count vector containing, for each topic, the number of words sampled from that topic in document $W_d$. That is $Z_d = \sum_{n=1}^{N_d} z_{dn}$. This is a latent/unobserved quantity. If $K$ and $\Phi$ were known and the $z_{dn}$'s were observed, then $Z_d$ would be a sufficient statistic in the sense that $p(W_d, (z_{dn})_{n = 1}^{N_d} \mid \Phi, \alpha)$ depends on $(z_{dn})_{n = 1}^{N_d}$ only through $Z_d$. For this reason we will use $p(W_d, Z_d \mid \Phi, \alpha)$ to denote $p(W_d, (z_{dn})_{n = 1}^{N_d} \mid \Phi, \alpha)$.
 
Pretend that $K$ and $\Phi$ are known to be equal to $\hat{K}$ and $\hat{\Phi}$, respectively. Now the data-likelihood
\[
    p(W_d \mid \Phi, \alpha) = \sum_{Z_d} p(W_d, Z_d \mid \Phi, \alpha) 
\]
is computationally intractable due to the combinatorial sum over $Z_d$. We will therefore be working with $p(W_d, Z_d \mid \Phi, \alpha)$. However, for this we will need to estimate the latent $Z_d$'s. We will estimate $z_{dn}$ by the canonical basis vector $e_{i^*} \in \R^{\hK}$, where $i^*$ is a topic that maximizes the posterior probability $p(w_{dn} \mid \hat{\phi}_i)$, that is
\[
   \hat{z}_{dn} = e_{i^*} \in \R^{\hat{K}}, \text{ with } i^* \in \arg \max_i p(w_{dn} \mid \hat{\phi}_i),
\]
where $p(w_{dn} \mid \hat{\phi}_i)$ is computed from the best LDA model obtained from Step 1. Then we can estimate $Z_d$ by
\[
    \hat{Z}_d = \sum_{n = 1}^{N_d} \hat{z}_{dn}
\]
and then work with $p(W_d, Z_d \mid \Phi, \alpha)$ as if $Z_d$ were known to be equal to $\hat{Z}_d$. Note that, $\widehat{Z}_d$ is a function of $W_d$.

For simplicity of notation we will subsequently use $K$, $\Phi$, $Z_d$ instead of $\hat{K}$, $\hat{\Phi}$ and $\hat{Z_d}$. The reader should keep in mind that their values are known (i.e. estimated).

The pseudocode for Sections \ref{sec:step1} and \ref{sec:step2} is presented in Algorithm \ref{alg:1}.

\subsection{Step 3: Changepoint analysis}
\subsubsection{Derivation of the changepoint statistic} \label{sec:step3a}
The joint likelihood of any document $W_d$ and its topic count vector $Z_d$ in $W$, given the topic vector $\Phi$ and the Dirichlet parameter $\alpha$, is given by
\begin{align} \nonumber
    &p(W_d, Z_d \mid \Phi,\alpha) \\ \nonumber
    &= \int p(\theta_d \mid \alpha)p(Z_d \mid \theta_d)p(W_d \mid Z_d,\Phi)d\theta_d \\ \label{eq:doc_likelihood}
    &= p(W_d \mid Z_d,\Phi)p(Z_d \mid \alpha).
\end{align}
Here $p(Z_d \mid \alpha)$ is the density of the P\'olya distribution (see, e.g., \cite{minka2000estimating}). Now we define our changepoint statistic $S(s,e)$ for the documents $W_s,\ldots,W_e$. We already have estimates of $\Phi$ and $Z_d$ from the previous subsection. In the following definition, $t = \frac{s + e}{2}$ and $T_{s, e}^{(\lr)}(t)$ is as defined in \eqref{eq:lr}: 
\begin{equation*}
    S(s, e) := \frac{1}{e - s + 1} \log T_{s, e}^{(\lr)}(t),
\end{equation*}
where the parameter of interest is $\theta = \alpha$ and the relevant density is $f_{\theta}(W_d,Z_d) = p(W_d,Z_d\mid\Phi,\alpha)$. Actually, due to the form of \eqref{eq:doc_likelihood}, $f_{\theta}(W_d,Z_d)$ in the definition of the LR statistic $T_{s, e}^{(\lr)}(t)$ may be replaced by $\tilde{f}_{\theta}(W_d, Z_d) = p(Z_d \mid \alpha)$. For a discussion on maximum likelihood estimation for P\'olya distribution, see Section~\ref{sec:polya}.

$S(s,e)$ gives a measure of distance between the model with exactly one changepoint at $t$ and the null model of no changepoints within $[s, e]$. Note that we are not computing the full LR statistic, we are only computing the LR statistic for the hypothesis that there is exactly one changepoint at $t = \frac{s + e}{2}$. Computing the full LR statistic is expensive because of the various maximizations involved. The idea behind $S(s, e)$ is that for small $[s, e]$, $S(s, e)$ should still capture the essence of the LR methodology while being computationally simpler.

\subsubsection{Threshold calculation} \label{sec:step3b}
For our statistic $S(s, e)$, we need to compute a threshold based on the null model of no changepoints in order to be able to detect changes. This threshold essentially measures the noise level when no signal is present. If our computed statistic $S(s, e)$ does not exceed the threshold, then we may be very confident that there is no changepoint within the interval $[s, e]$. On the other hand, if $S(s, e)$ exceeds the threshold, that provides some evidence regarding the existence of a changepoint. The larger the difference between $S(s, e)$ and the threshold, the stronger the evidence.

For computing the threshold, we need to know how our statistic would behave under a null model of no changepoints. Note here that the threshold should a priori depend on the Dirichlet parameter $\alpha$ under the null model. In fact, for statistics such as $S(s, e)$ or the CUSUM statistic which involve averages over time-indices in $[s, e]$, it is natural to expect, via a variance computation, that the noise level will be inversely proportional to square root of the length $e - s + 1$ of the interval. Thus our threshold should look like
\[
    g(\alpha) \times \frac{1}{\sqrt{e - s + 1}}.
\]
However, finding a good mathematical approximation to $g(\alpha)$ seems to be intractable. Thus we take empirical approaches, bypassing the estimation of $g(\alpha)$ and settling with possibly conservative estimates of it.

\textbf{Approach 1.}
As the documents are i.i.d. under the null model, we can use a permutation-testing type approach for computing a threshold. We take permutations of the documents in $[s, e]$ for different $s, e$ such that the length $e - s + 1$ is fixed. We use the following type of permutations: $W_s^*, \ldots W_e^*$, where $ W_{s + 2k} = W^*_{s+k}, \text{ and } W_{s + 2k + 1} = W^*_{t + k}$ with $t = \frac{s + e}{2}$.

Observe that, for the above permutation, the sequence of documents to the left of $t = \frac{s + e}{2}$ has the same distribution as the sequence to the right. If the interval $[s, e]$ actually contains a changepoint, then the statistic $S(s, e)$ computed on the permuted documents will not be distributed as under the null model. However, if we do this for many intervals of the same length $e - s + 1$, and there are not a lot of changepoints, then a majority of these intervals will have no changepoints. Therefore we can take a suitable quantile of the calculated statistics (e.g., the median) as the threshold. Our empirical results show good performance of this method. If we use largest empirical quantile, the threshold we get will be more conservative, requiring high signals for detection. In our real data examples, signals were high enough to pass even this conservative threshold. 

\textbf{Approach 2.}
Since, in our first step, we use LDA to estimate the topic polytope $\Phi$ and in the process get an estimate of $\alpha$, we can use that $\alpha$ and the estimated $\Phi$ to simulate documents from a topic model with no changes. Then, from this simulated corpus, we can estimate thresholds. Note, however, that the estimated $\alpha$ is only an average measure. Thus the thresholds we get from this approach may potentially have biases in regard to the $g(\alpha)$ term in the threshold.

\begin{figure}[!h]
\centering
\includegraphics[scale = 0.9]{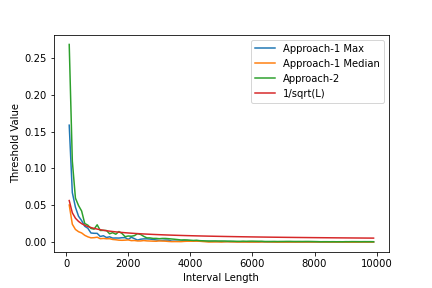}
\caption{Comparison of thresholds computed using different approaches.}
\label{fig:comp}
\end{figure}

We see empirically that Approach 1 with the maximum empirical quantile and Approach 2 produce similar thresholds. Approach 1 with median produces lower thresholds for small intervals. All the approaches agree with the expected $1/\sqrt{L}$ behaviour for moderate to large $L$ (here $L$ represents the length of intervals $[s, e]$ on which our changepoint statistics are computed). See Figure~\ref{fig:comp} for a comparison of these thresholds. The pseudocode for Sections \ref{sec:step3a} and \ref{sec:step3b} (with Approach 1) is presented in Algorithm \ref{alg:2}.

\subsubsection{Estimation of changepoints} \label{sec:step4}
To estimate changepoints, we use ideas from the WBS algorithm described earlier. However, our algorithm differs from traditional WBS in several key ways. Since WBS is designed for standard multivariate data, the threshold it uses is generic and not suitable for our purposes. Also, unlike traditional WBS, where there is only a single global threshold, we use a size-dependent threshold for all intervals of a fixed size. Hence checking if a score is below the corresponding threshold must be done before finding the maximum score, because, otherwise, it could so happen that the maximum score lies below its corresponding threshold, while there exist values less than the maximum which are greater than their corresponding thresholds, so that we terminate incorrectly. The pseudocode for this is presented in Algorithm \ref{alg:3}.

We compute the statistic values (scores) for all the sampled intervals and keep only those below their respective thresholds before we start recursively looking for changepoints using these filtered scores. Since the statistic values are computed using iterative gradient based methods, computing them beforehand saves a lot of time, for otherwise, one would have to compute the same statistic multiple times during the recursive calls. 

\section{Algorithms} \label{sec:algo}
In this section, we present pseudocodes of all our main algorithms.  Algorithm \ref{alg:1} estimates the underlying topic vectors and also prepares the topic count vectors $Z_d$ as described in Sections \ref{sec:step1} and \ref{sec:step2}. Algorithm \ref{alg:2} computes thresholds for different interval-lengths $\ell$ as described in Sections \ref{sec:step3a} and \ref{sec:step3b}.
Algorithm \ref{alg:3} is our modified version of WBS as described in Section \ref{sec:step4}. Algorithm \ref{alg:4}, then combines the previous algorithms and detects/estimates the changepoints. 

\begin{algorithm*}
\SetAlgoLined
\SetKwInOut{Input}{Input}
\SetKwInOut{Output}{Output}
\Input{$3$-way split $(\widetilde{W}_1, \widetilde{W}_2, W)$ of $W^{\full}$}
\Output{$Z$ (matrix of topic counts per document in $W$)}

Fit LDA models with different number of topics on $\widetilde{W}_1$, and report the model with the highest log-perplexity on $\widetilde{W}_2$. Let \LearnedTopicModel{} denote this model \\
$\Phi \leftarrow [\phi_1:\cdots:\phi_K]$, the $K$ topic vectors of \LearnedTopicModel{} \\
Let $P(w \mid \phi_j)$ denote the probability of a word $w$ coming from topic $j$ under \LearnedTopicModel{} \\
\For{$d = 1$ \KwTo $T$}{
    \For{$n = 1$ \KwTo $N_d$}{
        $i \leftarrow \arg \max_i P(w_{dn} \mid \phi_k)$ \\
        $Z[d, ] \leftarrow Z[d, ] + e_i$
    }
}
\caption{\EstimateTopicCounts{}}
\label{alg:1}
\end{algorithm*}

\begin{algorithm*}
\SetAlgoLined
\SetKwInOut{Input}{Input}
\SetKwInOut{Output}{Output}
\Input{Minimum interval size ($\delta$), topic counts ($Z$), quantile parameter $(\eta)$}
\Output{\Threshold{} (thresholds for all interval sizes)}
\tcp{$\ell$ denotes interval size}
\For {$\ell =  \delta$ \KwTo $T$} {
    $\intervals{} \leftarrow$ a list of $M$ random intervals of the form $[s_m, s_m + \ell -1]$ \\ 
    Initialize an empty list \scores{} \\
    \ForAll {$[s_m, e_m]$ in \intervals{}} {
        $t_m \leftarrow \frac{s_m + e_m}{2}$ \\
        Permute the rows of $Z$ to get $Z^*$ such that $\forall $ valid $k$,\\
        \qquad$Z_{s_m + 2k} = Z^*_{s_m + k}$ \\
        \qquad$Z_{s_m + 2k + 1} = Z^*_{t_m + k}$ \\
        Compute statistic $S(s_m, e_m)$ on $Z^*$ and append to \scores{}
    }
    $\Threshold{}[\ell] \leftarrow \eta$-th quantile of \scores{}
}
\caption{\CalculateThresholds{}}
\label{alg:2}
\end{algorithm*}

\begin{algorithm*}
\SetKwInOut{Input}{Input}
\SetKwInOut{Output}{Output}
\Input{$s$, $e$, \scores{}, \intervals{}, \changepoints{}}
\Output{\changepoints{}}
$M \leftarrow$ the number of elements in \scores{} \\
\If {$s = e$ \textbf{or} $M = 0$} {
    Terminate
}
$s_c, e_c \leftarrow \intervals{}[\arg\max_i\scores{}[i]]$ \\
$t_c \leftarrow \frac{s_c + e_c}{2}$ \\
Append $t_c$ to global list \changepoints{} \\
Initialize empty lists \preChangeIntervals{}, \postChangeIntervals{},
\preChangeScores{}, \postChangeScores{} \\
\For {$i = 1$ \KwTo $M$} {
    $s, e \leftarrow \intervals{}[i]$ \\
    \If {$e \leq t_c$} {
        append $[s, e]$ to \preChangeIntervals{} \\
        append $\scores{}[i]$ to \preChangeScores{}
    } \ElseIf {$s \geq t_c$} {
        append $[s, e]$ to \postChangeIntervals{} \\
        append $\scores{}[i]$ to \postChangeScores{}
    }
}
\mwbs{}($s$, $t_c$, \preChangeScores{}, \preChangeIntervals{}, \changepoints{}) \\
\mwbs{}($t_c$, $e$, \postChangeScores{}, \postChangeIntervals{}, \changepoints{})
\caption{\mwbs{}}
\label{alg:3}
\end{algorithm*}

\begin{algorithm*}
\SetAlgoLined
\SetKwInOut{Input}{Input}
\SetKwInOut{Output}{Output}
\Input{$W^{\full}$ (corpus), $\delta$ (minimum interval size), $\eta$ (quantile parameter)}
\Output{\changepoints{}}
$(\widetilde{W}_1, \widetilde{W}_2, W) \leftarrow$ $3$-way split of $W^{\full}$ \\
$Z \leftarrow \EstimateTopicCounts{}(\widetilde{W}_1, \widetilde{W}_2, W)$ \\
$\Threshold{} \leftarrow \CalculateThresholds{}(\delta, T, Z, \eta)$ \\
Initialize empty lists \scores{}, \intervals{}, \changepoints{} \\
\tcp{We sample $M$ random intervals}
\For{$i = 1$ \KwTo $M$}{
    Sample $[s, e]$, such that $e > s$ \\
    $\statistic{} \leftarrow S(s,e)$ \\
    \If {\statistic{} $\geq \Threshold{}[e - s + 1]$} {
        append \statistic{} to \scores{} \\
        append $[s, e]$ to \intervals{}
    }
}
$\changepoints{} \leftarrow \mwbs{}(1, T, \scores{}, \intervals{}, \changepoints{})$
\caption{\EstimateChangepoints{}}
\label{alg:4}
\end{algorithm*}

\section{Experiments} \label{sec:exp}
\subsection{Experiments on synthetic data} \label{sec:simu}
We first investigate the behaviour of our algorithm on synthetic data sets with different parameter configurations. The performance of our method is measured by comparing the estimated changepoints with the true changepoints using \textit{Precision}, \textit{Recall} and the \textit{$F$-score}. We count an estimated change as a correct one if it is within a tolerance window of the true changepoint (for this set of experiments we set the window to 50 documents). Precision is calculated by dividing the number of true positives by the total number of estimated changes, while Recall is computed by dividing the number of true positives by the number of true changes. The $F$-score is defined as $\frac{2 \times \text{Precision} \times \text{Recall}}{\text{Precision} + \text{Recall}}$.

\begin{table}[!ht]
\centering
\begin{tabular}{l|llll}
    \toprule
    \textbf{L} & \textbf{P} & \textbf{R} & \textbf{F} & $\widehat{K}$ \\
    \midrule
    0.1        & 0.89       & 0.93       & 0.85       & 16.8          \\
    0.3        & 0.96       & 0.97       & 0.94       & 13.1          \\
    1.0        & 0.99       & 0.98       & 0.98       & 13.6          \\
    3.0        & 0.99       & 0.99       & 0.99       & 7.1           \\
    \bottomrule
\end{tabular}
\caption{Precision (\textbf{P}), Recall (\textbf{R}), $F$-score (\textbf{F})  and the estimated number of topics ($\widehat{K}$) for experiments on synthetic data sets, averaged over 10 simulations for each value of the $\ell_2$-norm (\textbf{L}) of the Dirichlet parameters for the topic proportions, i.e. $\alpha_{1}, \ldots, \alpha_{M + 1}$. We use $4$ different values of \textbf{L}. The true number of topics is $K = 10$, with vocabulary size $V = 5000$, the total number of documents $T = 30000$ and the total number of changepoints $M = 20$. The changepoints were chosen randomly with the differences between consecutive changepoints in the range $[500, 3000]$.} 
\label{tab:results}
\end{table}
Table~\ref{tab:results} demonstrates that our algorithm works well across a wide range of parameter values. When the Dirichlet parameters for the topic proportions are small (resp. large), the documents tend to be sparse (resp. dense) mixtures of the topics. Our method performs well across all these cases. Also, it is able to estimate the changepoints correctly with quite uneven spacings between consecutive changepoints.

Note that the true number of topics is $K = 10$, but the average estimated number of topics ($\widehat{K}$) in Table~\ref{tab:results} is not equal to 10 in any of the four cases and still our algorithm estimates the changepoints almost perfectly, both when $\widehat{K} > K$ and $\widehat{K} < K$, as long as $|\widehat{K} - K|$ is reasonably small. We conclude that a reasonably close estimate of $K$ is sufficient for good estimation of changepoints.

Recall that we estimate the topic vectors using the part $W' \cup W''$ of the full corpus. We empirically found that the topic polytope, i.e. the convex hull of the topic vectors, learned using the entire corpus is very close to the one learned using only a representative part of it (as our algorithm does). More details on this appear in the supplementary material.

We also picked up subsets of documents from the time-series, learned the topic vectors on them using LDA, and found that they can be represented quite well within the topic polytope that our algorithm had learned. This supports our claim that the entire time-series can be represented by a large collection of potential topics, and documents can then be represented as sparse mixtures of those topics.

\begin{figure*}[!h]
\centering
\begin{tabular}{c}
    \includegraphics[scale =0.9]{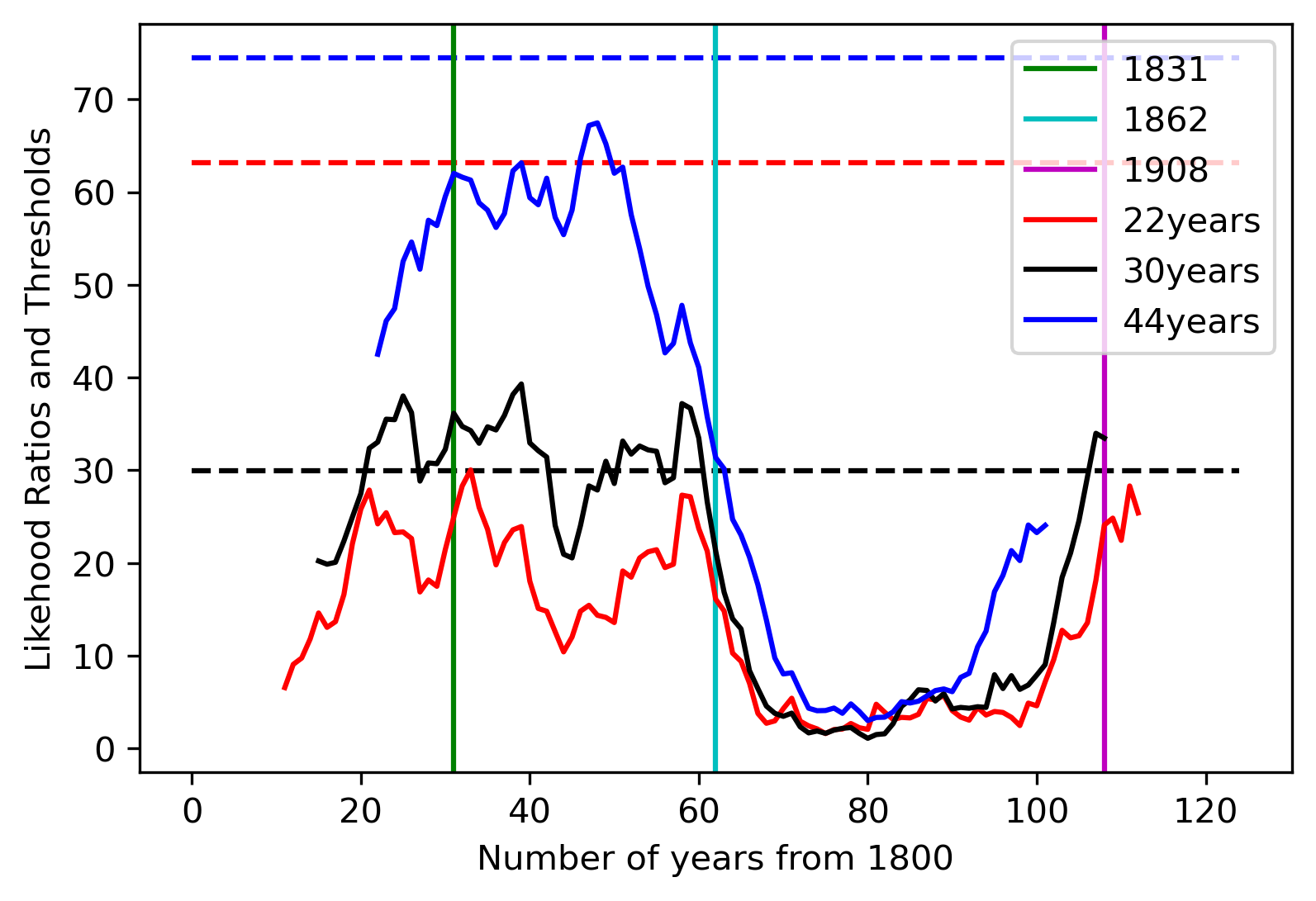} \\
    \includegraphics[scale = 0.9]{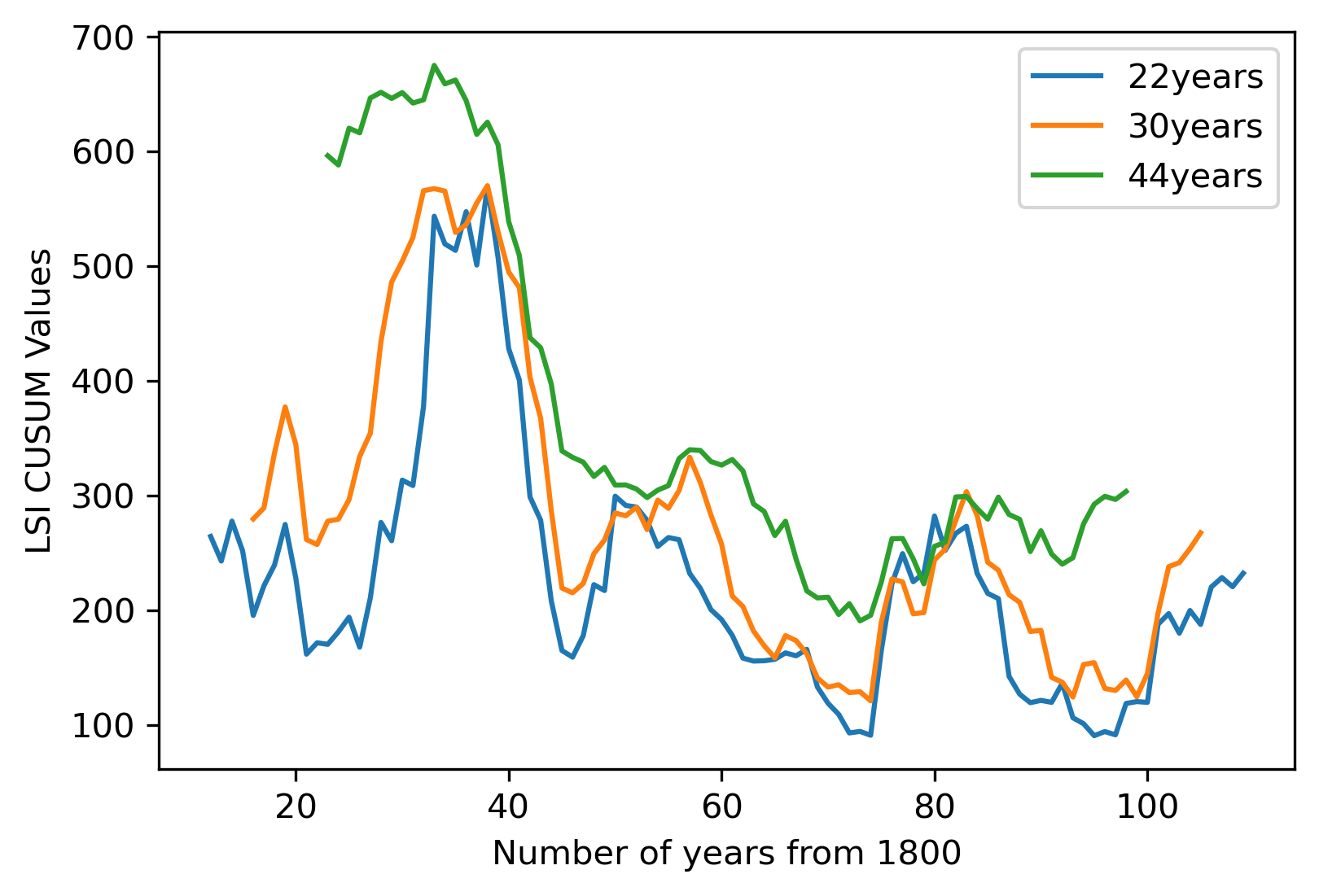}
\end{tabular}
\caption{The top plot shows our statistic $S$ and the bottom plot shows the multivariate CUSUM statistic applied on the LSA-based low-dimensional representations of the documents on all possible intervals of lengths 22, 30 and 44 years on the $y$-axis and the corresponding midpoint of each interval on the $x$-axis. The dashed horizontal lines are the respective thresholds, and the vertical lines mark the locations of the estimated changepoints.}
\label{fig:LR-LSI}
\end{figure*}

\subsection{Experiments on real data sets and comparisons with baselines} \label{sec:real-data}
We apply our method on two real data sets: (i) a corpus of English literature from the period 1800-1922 \citep{englit2015}; (ii) abstracts from the High Energy Physics arXiv repository\footnote{\url{https://www.kaggle.com/Cornell-University/arxiv}}
\citep{clement2019arxiv}.

\subsubsection{Baselines and alternatives}
We use the approach of \cite{wang2018real} as baseline, as their goal is very close to ours (the only difference being that they worked in online settings), and theirs is the only related work scalable to data sets of our size. We calculate cosine-similarities between the top-word probabilities of consecutive time-points and run univariate changepoint algorithms like PELT \citep{killick2012optimal} and BCP \citep{barry1993bayesian}, as done in \cite{wang2018real}. For BCP, we declare as significant the time-points which have a posterior probability $> 0.35$ of being a changepoint. 

We also compare against an alternative model-free approach based on LSA (which, to the best of our knowledge, has not been explored before in the literature). We obtain a $k$-dimensional representation of each document by using LSA (see Section~\ref{sec:lsa}) and use a multivariate CUSUM statistic on these representations along with the WBS algorithm.

\subsubsection{English literature}
This data set contains works in English literature published between the years 1700-1922 spanning fiction, poetry and drama. We work with works of fiction only. As there are too few documented works between 1700 and 1799, we focus on the period 1800-1922 only. As the number of documented works has increased substantially with time, we limit the number of works per year to a maximum of 150 so that each period gets a fair and uniform representation when we estimate the underlying topics. Thus we have 17645 works spread over 123 years. The data set comes with lists of most frequently occurring words per year recognized by an English dictionary. This helps us avoid proper nouns like names of characters in these works. Furthermore, we remove the common stop-words in English.The final vocabulary after this pre-processing has size $V = 5460$.

\begin{figure*}[!ht]
\centering
\begin{tabular}{c}
     \includegraphics[scale = 0.44]{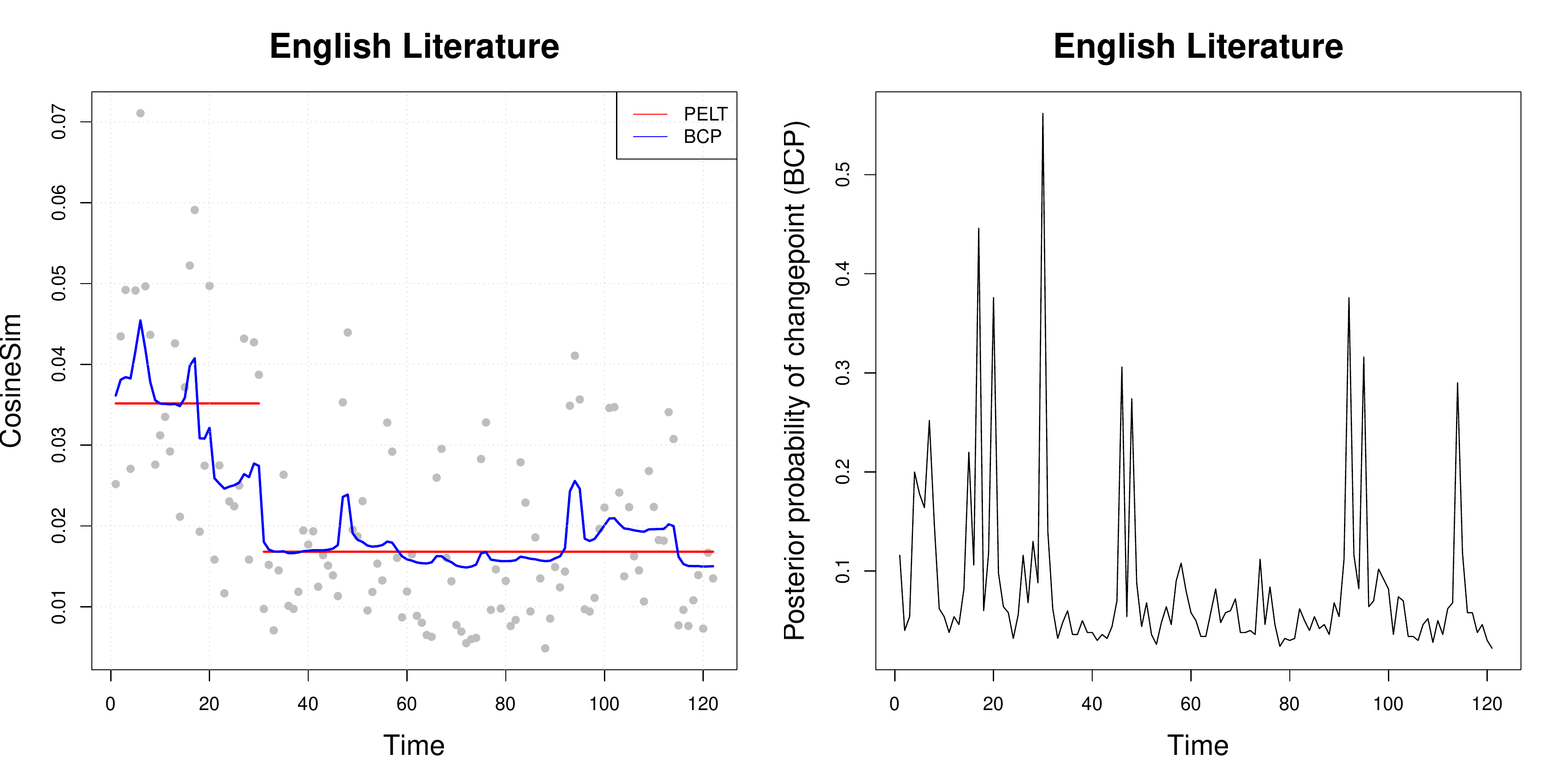} \\
     \includegraphics[scale = 0.44]{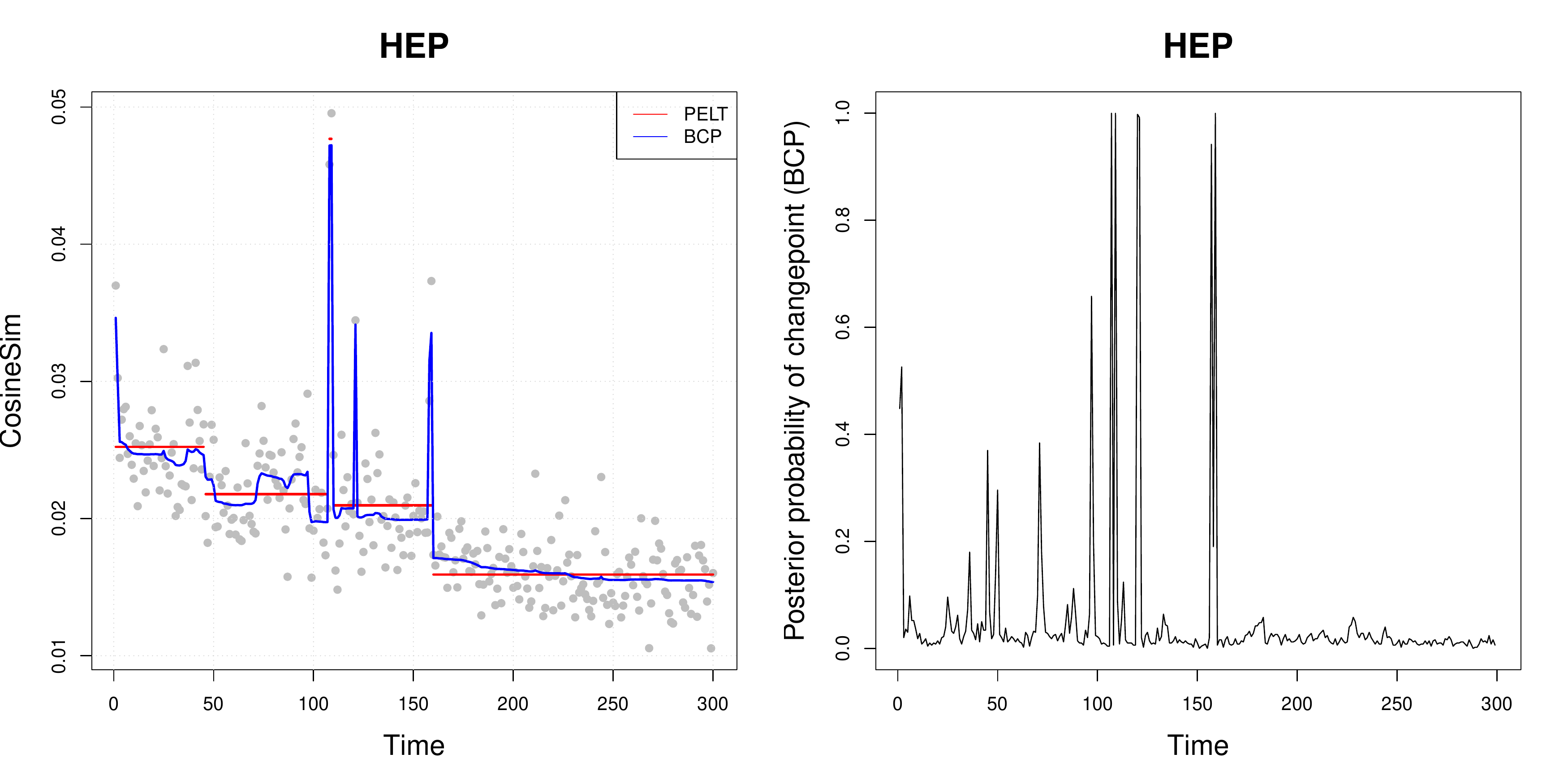}
\end{tabular}
\caption{Plots of the baseline changepoint algorithms for the English literature and HEP data sets.}
\label{fig:baseline}
\end{figure*}

Literature scholars (see \cite[pages 1, 289, 881]{robinson2018british}) demarcate 1789-1830 as the era of Romanticism and 1837-1901 as the Victorian era, with the Modern era beginning with WWI in the early 1910s. The Victorian era is often divided into the early Victorian and the late Victorian eras, with the change happening in the late 1860s. Results reported in Table \ref{tab:realdataexps}, show our estimated changepoints in agreement with literature scholars. Since the change near $t = 1830$ is a strong one marking the beginning of the Victorian era, all methods pick up this change. Being based on top-word probabilities which are local parameters unlike the global parameters of our model that undergo change, the baseline methods miss the other two changepoints. We also see that PELT is conservative in its estimation and hence prone to false negatives, whereas BCP is prone to false positives. See Figure~\ref{fig:baseline} for plots of PELT and BCP.

\begin{table*}[!ht]
{\footnotesize
\centering
\begin{tabular}{l|c|c}
    \toprule
    Method           & English literature data set & HEP data set (no. of months from August, 1995) \\
    \midrule
    PELT             & 1830                        & 45, 107, 159                                   \\
    BCP ($p > 0.35$) & 1817, 1820, 1830, 1892      & 1, 2, 45, 97, 107, 109, 120, 121, 157, 159     \\
    Ours             & 1831, 1862, 1908            & 36, 153, 192                                   \\
    \bottomrule
\end{tabular}
\caption{Changepoint estimates by the baselines and our method in real data sets.}
\label{tab:realdataexps}}
\end{table*}

\begin{figure}
\centering
\begin{tabular}{cc}
    \includegraphics[scale = 0.33]{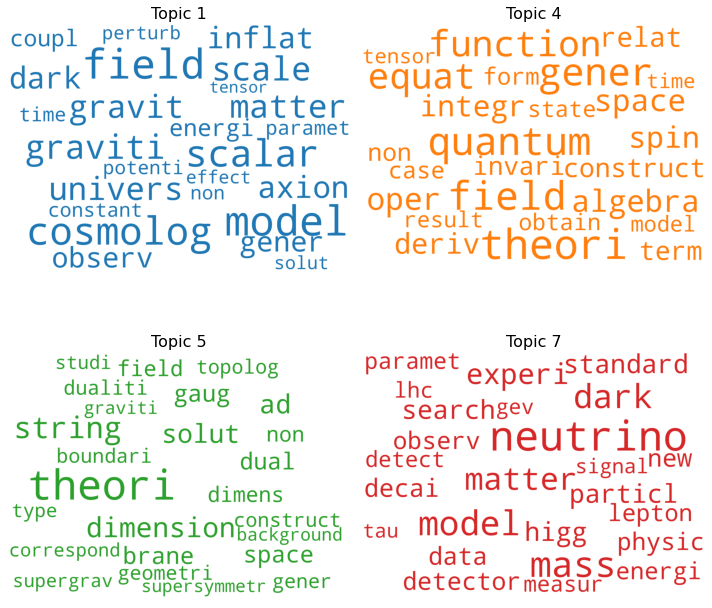} & \includegraphics[scale = 0.28]{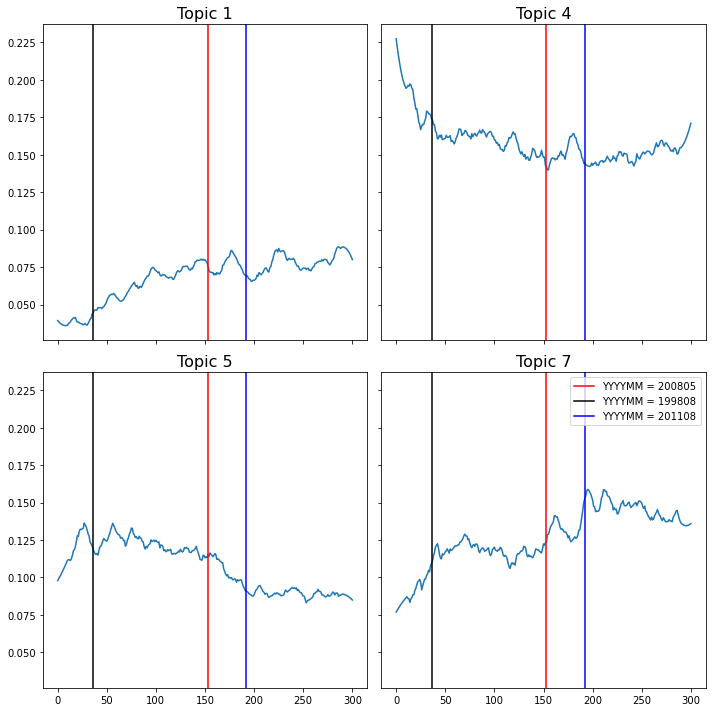}
\end{tabular}
\caption{Top words and monthly popularity for 4 topics from the HEP data set.}
\label{fig:HEP}
\end{figure}

Figure~\ref{fig:LR-LSI} demonstrates the advantages of our approach over the LSA based alternate approach. The statistic values under this approach have very high variability, and they hardly settle down to a constant value, as is expected in intervals with no change. For our topic model, we notice how the statistic values are constant (and below threshold) in the interval between 1862 and 1908. In case an interval contains more than one changepoint (e.g., the intervals of length 44 years in between 1831-1862), even when a peak exists in our statistic values at a wrong location due to possible interference of signals, the peak remains below the respective threshold, thus preventing false positives.

\subsubsection{High-Energy Physics abstracts from the arXiv}
This data set contains the abstracts of all papers submitted to the arXiv between April, 1986 and August, 2020, with one or more of the following tags: High Energy Physics - Experiment (hep-ex), High Energy Physics - Lattice (hep-lat), High Energy Physics - Phenomenology (hep-ph) and High Energy Physics - Theory (hep-th). As very few papers were submitted in the first few years, we work with the abstracts of papers from August, 1995 to August, 2020 only---a total of 281891 abstracts spread over 301 months. We again removed the common English stop-words and also retained only those words occurring more than 100 times over the entire corpus, thus giving us a vocabulary of $V = 4336$ words.

Our method estimated 3 changepoints: August, 1998 ($t = 36$), May, 2008 ($t = 153$) and August, 2011 ($t=192$). Figure~\ref{fig:HEP} displays the top words of the 4 topics whose popularities had significant changes over time, and how these popularities (measured in terms of monthly topic proportions) changed over the period of 301 months. To the best of our knowledge, changes in the field of HEP have not been studied rigorously before. So we now try to explain these detected changes. Topic 1 seems to represent Cosmology, which saw a lot of activity during the successful LIGO experiment. Topic 4 seems to be related to Quantum Field Theory. This field had some conclusive research activity in the last decade of the twentieth century, and is mostly used in applications nowadays. Topic 5 seems to be related to String Theory. It also peaked during the last decade of the twentieth century but slowed down due to the unavailability of conclusive experimental evidence. Finally, Topic 7 seems to be related to Particle Physics, which saw a conspicuous increase in research interest in the last decade owing to CERN's LHC experiments. In the period 2011-2013, concrete evidence for the existence of the Higgs Boson was found by the LHC, which led to a lot of excitement in this field. Our method predicts the year 2011 to be a changepoint, which agrees with the experiment years. As reported in Table \ref{tab:realdataexps}, both the baselines miss out on this changepoint near $t=192$. Once again, BCP is prone to many false positives. See Figure \ref{fig:baseline} for plots of PELT and BCP.

\subsection{Topic polytope estimation}
To evaluate our results we will be using the minimum-matching Euclidean distance which is defined as follows. Given two polytopes $\Phi$ and $\Phi'$, the distance is given by 
\begin{align*}
    d_M(\Phi, \Phi') = \max (d(\Phi, \Phi'), d(\Phi', \Phi)),
\end{align*}
with
\begin{align*}
    d(\Phi, \Phi') = \max_{\phi \in \mathrm{extr}(\Phi)}\min_{\phi' \in \mathrm{extr}(\Phi')} \|\phi - \phi'\|_2,
\end{align*}
where extr($\Phi$) and extr($\Phi'$) are the extremities of the polytopes $\Phi$ and $\Phi'$, respectively. Since, in our case, the number of words are much much larger than the number of topics, these extremities will be the topic vectors themselves (see, e.g., \cite{tang2014understanding}).

Also, when we refer to \textit{one sided error}, we mean $d(\Phi, \Phi')$. This captures the largest deviation of the topics in $\Phi$ from those in $\Phi'$.

\subsubsection{Distance between polytopes using full corpus and a part of the corpus}
As our algorithm uses only a part of the entire corpus for estimating the underlying topics, it is necessary to ensure that the convex hull of the learned topics are close to the convex hull of the topics that would have been learned if the entire corpus was used. We use the minimum matching Euclidean distance to measure the distance between the above mentioned convex hulls (polytopes).  

\begin{table}[h]
\centering
\begin{tabular}{l|ll}
    \textbf{L} & \textbf{D} & \textbf{OD} \\
    \hline
    0.1        & 0.104      & 0.059       \\
    0.3        & 0.063      & 0.051       \\
    1.0        & 0.043      & 0.040       \\
    3.0        & 0.034      & 0.029       \\
    \hline
\end{tabular}
\caption{\textbf{D} = average minimum-matching distance between polytopes learned from the entire corpus ($W^{\full}$) and the part used for training ($\widetilde{W}_1$); \textbf{OD} = one-sided error between 1000 arbitrary adjacent documents and the learned polytope, Both measures are averaged over 10 i.i.d. corpora for each shown $\ell_2$-norm value (\textbf{L}) of the Dirichlet parameter $\alpha$.}
\label{tab:min-matching}
\end{table}

\subsubsection{Arbitrary documents and the learned topic polytope}
To test if an arbitrary sequence of documents can be represented on the learned topic polytope, we select 10 random intervals $S$, where we learn the underlying topics for documents in the interval for each $[s_m, e_m] \in S$, and then compute the one-sided error between these learned topics and the learned topics by our algorithm. Even though such arbitrary intervals might have had only a few representative topics of the entire set of topics our corpus has, the results suggest that the topic polytope we learn is able to accommodate these topics fairly well.

For the High Energy Physics (HEP) data, we expect that the polytopes for HEP-EX (experimental) abstracts and HEP-TH (theoretical) abstracts should be quite different, but still we show that the one sided error for the topics estimated using the both the abstracts with respect to those estimated using just one of them individually is quite less---0.070 and 0.078 respectively---indicating that we can learn all the underlying topics together without compromising the representations of either of them.

\section{An LSA-based approach} \label{sec:lsa-based-method}
\begin{figure}[!h]
\centering
\includegraphics[scale = 0.75]{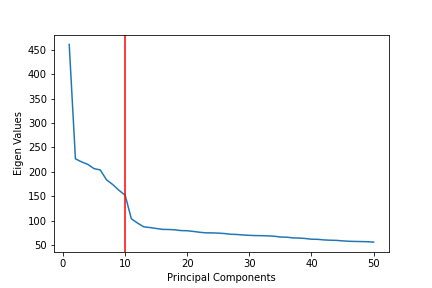} 
\caption{Scree plot to choose the number of components in LSA (the true number of topics is 10)}
\label{fig:scree}
\end{figure}

\begin{figure}[!h]
\centering
\includegraphics[scale = 0.75]{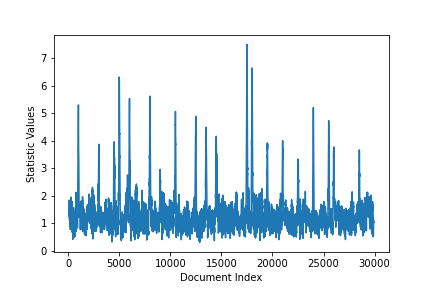} 
\caption{ Plot of the $\ell_2$-norm of the CUSUM statistic as a function of time (i.e. document index) on a simulated data set with $20$ true changepoints.}
\label{fig:LSA}
\end{figure}

Our approach to dimensionality reduction is via topic models. An alternative approach would be to use Latent Semantic Analysis (LSA) to get low-dimensional representations of the documents and then apply an off-the-shelf multivariate changepoint detection procedure on these low-dimensional representations. Using a Scree plot of the singular values we can decide on the dimension to choose as demonstrated in Figure~\ref{fig:scree}. In Figure~\ref{fig:LSA}, we plot the $\ell_2$-norm of a vector CUSUM statistic (defined in Section \ref{sec:prelim}) computed on low-dimensional representations obtained from LSA as a function of time (document index). Although we get peaks near all the true changepoints, note the high variability of the statistic. Also, unlike the topic modeling approach, the LSA based approach does not offer easy interpretable of the obtained changepoints.

\section{Conclusion} \label{sec:conc}
In conclusion, we have proposed in this article an offline method for changepoint analysis in textual data that utilises topic-modelling to reduce dimensionality and enable efficient detection and estimation of changepoints. Currently, most text corpora have temporally inhomogeneous volumes of text which makes it challenging to learn parameters for old times. As we move further into the 21-st century, our text-corpora are bound to grow, leaving us with a gold-mine of offline data that could be automatically analysed for structural changes, and this necessitates development of scalable methods that could handle millions of large documents with ease. We believe that the present article takes a small step towards that goal. 

\bibliographystyle{chicago}
\bibliography{main}

\newpage
\appendix

\section{P\'olya distribution and ML estimation} \label{sec:polya}
The Dirichlet-multinomial distribution is a compound distribution where $\theta$ is drawn from a Dirichlet distribution with parameter $\alpha = (\alpha_1,\ldots,\alpha_K)^\top$ and then a sample of discrete outcomes $z$ is drawn from a multinomial with probability vector $\theta$ (this $z$ is basically our topic count vector $Z_d$ for a document $d$). This compounding is essentially a P\'olya urn scheme, so the Dirichlet-multinomial is also called the P\'olya distribution. Let $n_k$ be the number of times the outcome was $k$. The resulting distribution over $z$, a vector of discrete outcomes, is given by
\begin{align*}
    p(z\mid\alpha) &= \int p(z\mid\theta)p(\theta\mid\alpha)\, d\theta \\
               &= \frac{\Gamma(\sum_k \alpha_k)}{\Gamma(\sum_k n_k + \alpha_k)}\prod_k \frac{\Gamma(n_k + \alpha_k)}{\Gamma(\alpha_k)}, 
\end{align*}
where $\Gamma$ denotes the Gamma function.

Given a set of $N$ count vectors, $Z = \{z_1,\ldots,z_N\}$, we need to find the maximum likelihood estimate of the parameter $\alpha$ from the likelihood
\begin{align*}
    p(Z\mid\alpha) = \prod_k p(z_k\mid\alpha).
\end{align*}
Unfortunately, no closed form expression for a maximum likelihood estimate of $\alpha$ is available. Thus we have to resort to iterative gradient based methods. We provide here, the gradient of the log likelihood of $p(Z\mid\alpha)$:
\begin{align*}
    \frac{d\log(p(Z\mid\alpha))}{d\alpha_k} = \sum_i\bigg(\Psi\big(\sum_k \alpha_k\big) - \Psi\big(n_i + \sum_k \alpha_k\big) + \Psi(n_{ik} + \alpha_k)- \Psi(\alpha_k)\bigg),
\end{align*}
where $\Psi = \Gamma'/\Gamma$ is the digamma function.

\section{Some experimental details} \label{sec:more-exp-details}
\subsection{Choice of Dirichlet parameters while generating synthetic data sets}
While generating a sequence of documents consisting $M$ changepoints ($\tau_1,\ldots,\tau_M$), we need $M + 1$ different values of the Dirichlet parameter, i.e. $\alpha_1, \ldots, \alpha_{M+1}$. Documents corresponding to the $i$-th interval (i.e. $[\tau_{i-1}, \tau_i]$) are generated from $\dirich{(\alpha_i)}$. For any particular sequence, the $\ell_2$-norms of all these Dirichlet parameters ($\|\alpha_1\|_2, \ldots, \|\alpha_M\|_2$) were equal. We tried 4 different values of this equal norm value: 0.1, 0.3, 1 and 3. For each such value, we generated 10 sequences and reported the average scores. For any particular sequence, in order to ensure that the $i$-th interval and the $(i + 1)$-th interval have different generating distributions, we ensured that $\frac{\|\alpha_{i+1} - \alpha_{i}\|_2}{\|\alpha_i\|_2} \geq \epsilon$, where $\epsilon$ is a hyperparameter, which controls how far apart the parameters of consecutive intervals are. For all our reported values, we used the value $\epsilon = 0.5$.

\subsection{Hyperparameters for experiments on real data sets}
We had two hyperparameters during our experiments on real data sets. The minimum interval length for which we computed the statistics ($\delta$) and the overall number of intervals we considered ($N$). For both data sets we set $N$ to be 5 times the time-series length. So for the English literature data set, we had about 500 intervals in total of varying sizes and for the HEP data set, we had about 1500 intervals in total of varying sizes. The value of $\delta$ was set to 20 years for the English literature data set and to 12 months for the HEP data set.

\end{document}